\documentclass{article}
\usepackage{spconf,amsmath,graphicx,subfigure}
\usepackage{booktabs}

\usepackage{multirow}
\usepackage{array}
\usepackage{hyperref}
\usepackage{algorithm}
\usepackage{algorithmic}
\usepackage{makecell}

\title{Multi-Plateau Ensemble for Endoscopic Artefact Segmentation and Detection}

\name{Suyog Jadhav,
Udbhav Bamba,
Arnav Chavan,
Rishabh Tiwari,
Aryan Raj
}
\address{Indian Institute of Technology (ISM), Dhanbad}

\begin{document}
\maketitle
\thispagestyle{empty}
\pagestyle{empty}
\begin{abstract}
Endoscopic artefact detection challenge consists of 1) Artefact detection, 2) Semantic segmentation, and 3) Out-of-sample generalisation. For Semantic segmentation task, we propose a multi-plateau ensemble of FPN\cite{lin2016feature} (Feature Pyramid Network) with EfficientNet\cite{tan2019efficientnet} as feature extractor/encoder. For Object detection task, we used a three model ensemble of RetinaNet\cite{lin2017focal} with Resnet50\cite{he2015deep} Backbone and FasterRCNN\cite{ren2015faster} (FPN + DC5\cite{dai2017deformable}) with Resnext101 Backbone\cite{xie2016aggregated, hu2017squeezeandexcitation}. A PyTorch implementation to our approach to the problem is available at \href{https://github.com/ubamba98/EAD2020}{github.com/ubamba98/EAD2020}. 

\textbf{\emph{Index Terms-}} Endoscopy, FPN, EfficientNet, RetinaNet, Faster RCNN, Artefact detection.

\end{abstract}

\section{Datasets}
The given dataset of EndoCV-2020 \cite{EAD2019endoscopyDatasetI, EAD2019endoscopyDatasetII, ALI2020:SREP} has a total of 643 images for the segmentation task which we divided into three parts - train (474 images), validation (99 images) and holdout (70 images) in the sequence they were released. We made sure that the distribution of train and holdout were similar and that of validation was different. Validation set ensured that the model was not overfitting to the training data and at the same time generalizing well on holdout. For Detection, a similar strategy was adopted - train (2200 images), validation (232 images) and holdout (99 images)

\section{Methods}
\subsection{Data Pre-processing and Augmentations}
Due to the variable aspect ratios and sizes in the training data, we adopted a stage-dependent rescaling policy. During the training stage, we cropped the images to a fixed size of 512x512 without resizing. This made sure that the spatial information was not lost and at the same time, the input to models was within trainable limits. During validation and testing time, we padded the images such that both dimensions are a multiple of 128 which is required for the EfficientNet backbone (to handle max-pooling in deeper models). As the number of samples in the dataset were relatively low and unbalanced, various augmentation techniques were adopted to prevent overfitting and achieve generalization. Horizontal and vertical flip, cutout (random holes)\cite{devries2017improved}, random contrast, gamma, brightness, rotation were tested out. To strongly regularize the data we propose the use of CutMix for segmentation (Algorithm: \ref{algo:CutMix}) which was toggled on and off depending upon the variance of model outputs.

\begin{algorithm}
\caption{CutMix for Segmentation}
\label{algo:CutMix} 
\begin{algorithmic}
\FOR{each iteration}
\STATE input, target = get minibatch(dataset)
\IF{mode == training}
\STATE input s, target s = shuffle minibatch(input, target)
\STATE lambda = Unif(0,1)
\STATE r\_x = Unif(0,W)
\STATE r\_y = Unif(0,H)
\STATE r\_w = Sqrt(1 - lambda)
\STATE r\_h = Sqrt(1 - lambda)
\STATE x1 = Round(Clip(r\_x - r\_w / 2, min=0))
\STATE x2 = Round(Clip(r\_x + r\_w / 2, max=W))
\STATE y1 = Round(Clip(r\_y - r\_h / 2, min=0))
\STATE y2 = Round(Clip(r\_y + r\_h / 2, min=H))
\STATE input[:, :, x1:x2, y1:y2] = input\_s[:, :, x1:x2, y1:y2]
\STATE target[:, :, x1:x2, y1:y2] = target\_s[:, :, x1:x2, y1:y2]
\ENDIF
\STATE output = model\_forward(input)
\STATE loss = compute\_loss(output, target)
\STATE model\_update()
\ENDFOR
\end{algorithmic}
\end{algorithm}

For Object detection spatial transformations - flip and random scaling and rotating were used.

\begin{table*}[ht!]
\caption[]{Multi-Plateau Results}
\label{table:Multi-Plateau Results} 
\centering
\begin{tabular}{@{}|c|c|l|c|c|@{}}
\hline
\textbf{Encoder}         & \textbf{Optimizer} & \textbf{Loss function}    & \textbf{Validation (DICE)} & \textbf{FineTuning including Holdout} \\ [0.5ex] \hline
                &           & DICE             & \textbf{0.4917}            & 0.4900                               \\ 
                &           & BCE+DICE         & 0.4382            &  --                                  \\ 
                & Ranger    & BCE              & 0.4415            &  --                                 \\ 
                &           & BCE+DICE+JACCARD & \textbf{0.4771}   & 0.4630                              \\ [1.0ex]
Efficientnet B3 &           & DICE             & 0.4509            &    --                                \\ 
                &           & BCE+DICE         & 0.4500            &     --                               \\ 
                &  Over9000 & BCE              & 0.4170            &     --                               \\ 
                &           & BCE+DICE+JACCARD & 0.4525            &      --                              \\ [0.5ex] \hline
                &           & DICE             & 0.4568            &     --                               \\ 
                &            & BCE+DICE         & \textbf{0.4759}            & 0.4720                             \\ 
                &  Ranger   & BCE              & 0.4165            &     --                               \\ 
                &           & BCE+DICE+JACCARD & 0.\textbf{4718}            & 0.4666                              \\ [1.0ex]
Efficientnet B4 &           & DICE             & 0.3890             &   --                                 \\ 
                &           & BCE+DICE         & 0.4597            &    --                                \\ 
                & Over9000  & BCE              & 0.4151            &    --                                \\ 
                &           & BCE+DICE+JACCARD & 0.4614            &     --                               \\ [0.5ex] \hline
                &           & DICE             & \textbf{0.4761}            & \textbf{0.4987}                              \\ 
                &           & BCE+DICE         & 0.4693            & 0.4643                              \\ 
                &  Ranger   & BCE              & 0.4374            &  --                                  \\ 
                &           & BCE+DICE+JACCARD & \textbf{0.4781}            & 0.4900                              \\ [1.0ex]
Efficientnet B5 &           & DICE             & 0.4352            &   --                                                 \\ 
                &           & BCE+DICE         & \textbf{0.4730}             & 0.4823                              \\ 
                &  Over9000 & BCE              & 0.4151            &  --                                  \\ 
                &           & BCE+DICE+JACCARD & \textbf{0.4726}            & 0.4798                             \\ [0.5ex] \hline
\end{tabular}
\end{table*} 

\subsection{Multi-Plateau Approach}
Due to high variability and early overfitting nature in the dataset, the main focus was on making a strong ensemble by training models on different optimisation plateaus. A total of 8 different plateaus with permutations of two different optimisers and four different loss functions were optimised with EfficientNet backbone increasing the depth, width and resolution three times, going from B3 to B5 (Table \ref{table:Multi-Plateau Results}). For optimisers, Ranger and Over9000 were used. Ranger is a synergistic optimiser combining RAdam (rectified Adam)\cite{liu2019variance} and LookAhead\cite{zhang2019lookahead}, and Over9000 is a combination of Ralamb\cite{you2017large} and LookAhead. A total of 2*4*3 = 24 models were trained, but in the final ensemble, only the models with a dice greater than 0.47 were considered. Average pixel-wise ensembling was adopted.

\subsection{Multi Stage Training}
Complete segmentation training pipeline was divided into four stages -\\
Stage 1 - CutMix was disabled to reduce regularization effect, encoder was loaded with ImageNet weights and freezed for the decoder to learn spatial features without being stuck into saddle point, crops were taken with at least one pixel having a positive mask. \\
Stage 2 - CutMix was enabled for strong regularization, and encoder was unfreezed to learn spatial features of endoscopic images. \\
Stage 3 - Random crops were trained instead of non-empty crops for the model to learn negative samples. \\
Stage 4 - Very few epochs with CutMix disabled and encoder freezed for generalization on original data. \\
For every consecutive stage best checkpoint of the previous stage was loaded. 

\subsection{Triple Threshold}
After analysing predictions on holdout, it was found that the number of false positives was quite high. To counter this, we implemented a novel post-processing algorithm which specifically reduced the number of false positives in the predictions (Algorithm \ref{algo:Triple Threshold}). Three sets of thresholds - max\_prob\_thresh, min\_prob\_thresh, min\_area\_thresh were tuned for this given task.

\begin{algorithm}
\caption{Triple Threshold}
\label{algo:Triple Threshold} 
\begin{algorithmic}
\FOR{each\_sample}
\STATE output\_masks = model(each\_sample)
\STATE final\_masks = []
\STATE i = 0
\FOR{each output\_mask in output\_masks}
\STATE max\_mask = output\_mask $>$  max\_prob\_thresh\
\IF{max\_mask.sum() $<$ min\_area\_thresh[i]}
\STATE output\_mask = zeros(output\_mask.shape)
\ELSE
\STATE output\_mask = output\_mask $>$ min\_prob\_thresh\
\ENDIF
\STATE i = i + 1
\STATE final\_masks.append(output\_mask)
\ENDFOR
\ENDFOR
\end{algorithmic}
\end{algorithm}

max\_prob\_thresh and min\_prob\_thresh were tuned using grid search on holdout dataset, whereas min\_area\_thresh was calculated by sorting sum of positive pixels of every class and taking the $2.5^{th}$ percent respectively. min\_area\_thresh used after calculation were 2000, 128, 256, 256 and 1024 respectively for every class. The results for triple threshold on single best model are compiled in Table \ref{table:Triple Threshold Results1} and comparison of our best performing models in Table \ref{table:Triple Threshold Results2}.
 
\begin{table*}[ht!]
\caption[]{Object Detection Ensembling}
\label{table:Object Detection Ensembling} 
\centering
\begin{tabular}{@{}|c|c|c|@{}}
\hline
\textbf{Parameter} & \textbf{Values Tested} & \textbf{Description} \\ [0.5ex] \hline
                        &                   &   If two overlapping boxes have \\
      iou\_thresh       & 0.4, 0.5, 0.6     &   IoU value $>$ iou\_thresh, \\
                        &                   &   one of the boxes is rejected.\\ \hline
                        &                   &   If a predicted box has a \\
      score\_thresh     & 0.4, 0.5, 0.6     &   confidence value $<$ score\_thresh \\
                        &                   &   associated with it, the box is rejected.\\ \hline
                        &                                  & The weights are given to the predictions \\
                        & [1, 1, 1], [1, 1, 2], [1, 2, 1], & by each of the models. \\
      weights           & [2, 1, 1], [1, 2, 2], [2, 1, 2], & A model with higher weight has more \\
                        & [2, 2, 1]                        & influence on the final output than \\
                        &                                  & a model with a lower weight.\\ \hline
\end{tabular}
\end{table*} 

\begin{table}[ht!]
  \begin{center}
    \caption{Triple Threshold Results}
    \label{table:Triple Threshold Results1} 
    \begin{tabular}{l|c|r} 
      \textbf{Min Thresh} & \textbf{Max Thresh} & \textbf{Val Precision}\\
      \hline
      0.5 & - & 0.597\\
      0.5 & 0.6 & \textbf{0.601}\\
      0.5 & 0.7 & \textbf{0.608}\\
      0.5 & 0.8 & 0.598\\ \hline
      0.4 & - & 0.588\\
      0.4 & 0.6 & 0.593\\
      0.4 & 0.7 & \textbf{0.600}\\
      0.4 & 0.8 & 0.591
    \end{tabular}
  \end{center}
 " - " indicates no triple threshold
\end{table}

\begin{table}[ht!]
  \begin{center}
    \caption{Precision Values on Best Performing Models}
    \label{table:Triple Threshold Results2} 
    \begin{tabular}{l|c|r} 
      \textbf{Model} & \textbf{No Triple} & \textbf{Triple}\\
      \hline
      B3-Ranger-DICE & 0.492 & 0.494\\
      B5-Ranger-DICE & 0.597 & \textbf{0.608}\\
      B5-Ranger-BCE+DICE+JACCARD & 0.549 & \textbf{0.561}\\
      B5-Over9000-BCE+DICE & 0.520 & 0.530\\
      B5-Over9000-BCE+DICE+JACCARD & 0.505 & 0.515\\
    \end{tabular}
  \end{center}
\end{table}

\subsection{Object Detection}
For Object Detection, individual models were trained with SGD as optimizer and confidence threshold of 0.5. To counter the variance and improve the performance of our model predictions, general ensembling was performed. Retinanet with backbones of FPN and Resnet50 and Faster RCNN with backbones of FPN and Resnext 101 32xd were trained (Table \ref{table:Object Detection Results}). Our ensemble strategy involves finding overlapping boxes of the same class and average their positions while adding their confidences. For finding the best parameters for ensembling the three models’ predictions, we ran a grid search with all possible combinations of the given range of values (Table \ref{table:Object Detection Ensembling}).

\begin{table}[ht!]
  \begin{center}
    \caption{Object Detection Results}
    \label{table:Object Detection Results} 
    \begin{tabular}{l|c|r} 
      \textbf{Model} & \textbf{Val. mAP} & \textbf{Hold. mAP} \\ \hline
      RetinaNet (FPN backend) & 26.07 & 24.66\\
      Faster RCNN (FPN backend) & 20.11 & 21.47\\
      Faster RCNN (DC5 backend) & 27.64 & 26.15\\
      Ensembled & \textbf{32.33} & \textbf{30.12}\\
    \end{tabular}
  \end{center}
\end{table}

\section{Results}
We achieved a Segmentation score which was a weighted linear combination of dice, IOU and F2 of 0.5675 on the final leader board and for object detection task, an mAP of 0.2061 was obtained.

\section{Discussion \& Conclusion}
Gastric cancer accounts for around 1 million deaths each year which can be prevented by early diagnosis. In this paper we explored multi-plateau ensemble to generalize pixel level segmentation and localization of artefacts in endoscopic images. We developed novel augmentation and post-processing algorithms for better and robust model convergence. 
\newpage
\bibliography{ref.bib}{}
\bibliographystyle{unsrt}

\end{document}